\documentclass[11pt,a4paper]{article}
\usepackage[hyperref]{acl2020}
\usepackage{times}
\usepackage{latexsym}

\usepackage{microtype}

\aclfinalcopy 
\def\aclpaperid{8} 

\usepackage{graphicx}
\usepackage{multirow}
\usepackage{booktabs}
\usepackage{paralist}
\usepackage{subcaption}

\newcommand{\citesub}[1]{\citeauthor{#1} \shortcite{#1}}

\title{Is Your Goal-Oriented Dialog Model Performing Really Well?\\ Empirical Analysis of System-wise Evaluation}

\author{
 Ryuichi Takanobu$^1$, Qi Zhu$^1$, Jinchao Li$^2$, Baolin Peng$^2$, Jianfeng Gao$^2$, Minlie Huang$^1$\footnotemark[1]\\ 
 $^1$DCST, Institute for AI, BNRist, Tsinghua University, Beijing, China \\
 $^2$Microsoft Research, Redmond, USA\\
  $^1${\small \tt \{gxly19,zhu-q18\}@mails.tsinghua.edu.cn \quad aihuang@tsinghua.edu.cn} \\
  $^2${\small \tt \{jincli,bapeng,jfgao\}@microsoft.com}
}

\begin{document}
\maketitle
\renewcommand{\thefootnote}{\fnsymbol{footnote}}
\footnotetext[1]{Corresponding author}
\renewcommand{\thefootnote}{\arabic{footnote}}
\begin{abstract}
There is a growing interest in developing goal-oriented dialog systems 
which serve users in accomplishing complex tasks through multi-turn conversations. 
Although many methods are devised to 
evaluate and improve the performance of individual dialog components, 
there is a lack of comprehensive empirical study on how different components contribute to the overall performance of a dialog system. 
In this paper, we perform a system-wise evaluation and present an empirical analysis on different types of dialog systems which are composed of different modules in different settings. 
Our results show that (1) a pipeline dialog system trained using fine-grained supervision signals at different component levels often obtains better performance than the systems that use joint or end-to-end models trained on coarse-grained labels, (2) component-wise, single-turn evaluation results are not always consistent with the overall performance of a dialog system, and (3) despite the discrepancy between simulators and human users, simulated evaluation is still a valid alternative to the costly human evaluation especially in the early stage of development.
\end{abstract}

\section{Introduction}

\begin{figure}[!tb]
    \centering
    \includegraphics[width=\linewidth]{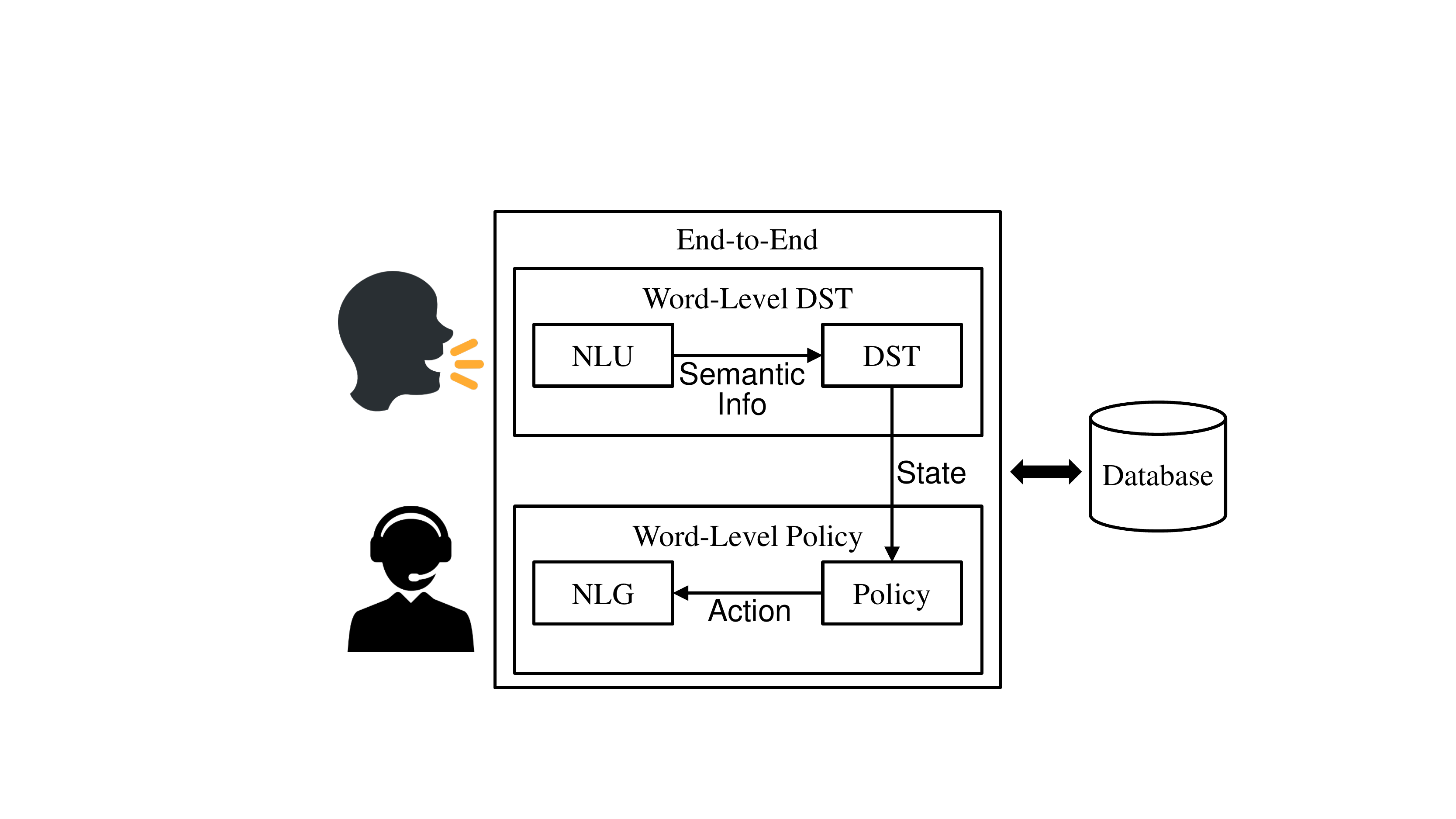}
    \caption{Different architectures of goal-oriented dialog systems. It can be constructed as a pipeline or end-to-end system with different granularity.}
    \label{fig:system}
\end{figure}

Many approaches and architectures have been proposed to develop goal-oriented dialog systems to help users accomplish various tasks \cite{gao2019neural,zhang2020recent}. 
Unlike open-domain dialog systems, which are designed to mimic human conversations rather than complete specific tasks and are often implemented as end-to-end systems, 
a goal-oriented dialog system has access to an external database on which to inquire about information to accomplish tasks for users. 
Goal-oriented dialog systems can be grouped into three classes based on their architectures, as illustrated in Fig.~\ref{fig:system}.
The first class is the pipeline (or modular) systems which typically consist of the four components: \textit{Natural Language Understanding} (NLU) \cite{goo2018slot,pentyala2019multi}, \textit{Dialog State Tracker} (DST) \cite{xie2015recurrent,lee2016task}, \textit{Dialog Policy} \cite{peng2017composite,takanobu2019guided}, and \textit{Natural Language Generation} (NLG) \cite{wen2015semantically,balakrishnan2019constrained}. 
The second class is the end-to-end (or unitary) systems \cite{williams2017hybrid,dhingra2017towards,liu2018dialogue,lei2018sequicity,qin2019entity,mehri2019structured}, which use a machine-learned neural model to generate a system response directly from a dialog history. 
The third one lies in between the above two types, where some systems use joint models that combine some (but not all) of the four dialog components. For example, a joint word-level DST model combines NLU and DST \cite{zhong2018global,wu2019transferable,gao2019dialog}, and a joint word-level policy model combines dialog policy and NLG \cite{chen2019semantically,zhao2019rethinking,budzianowski2019hello}. 

It is particularly challenging to properly evaluate and compare the overall performance of goal-oriented dialog systems due to the wide variety of system configurations and evaluation settings. Numerous approaches have been proposed to tackle different components in pipeline systems, whereas these modules are merely evaluated separately. Most studies only compare the proposed models with baselines of the same module, assuming that a set of good modules can always be assembled to build a good dialog system, but rarely evaluate the overall performance of a dialog system from the system perspective. A dialog system can be constructed via different combinations of these modules, but few studies investigated the overall performance of different combinations \cite{kim2019eighth,li2020results}. 
Although end-to-end systems are evaluated in a system-wise manner, none of such systems is compared with its pipeline counterpart. 
Furthermore, unlike the component-wise assessment, system-wise evaluation requires simulated users or human users to interact with the system to be evaluated via multi-turn conversations to complete tasks. 

To this end, we conduct both simulated and human evaluations on dialog systems with a wide variety of configurations and settings using a standardized dialog system platform, Convlab \cite{lee2019convlab}, on the MultiWOZ corpus \cite{budzianowski2018multiwoz}. 
Our work attempts to shed light on evaluating and comparing goal-oriented dialog systems by conducting a system-wise evaluation and a detailed empirical analysis. 
Specifically, we strive to answer the following research questions: 
(RQ1) Which configurations lead to better goal-oriented dialog systems? (\S \ref{sec:config}); 
(RQ2) Whether the component-wise, single-turn metrics are consistent with system-wise, multi-turn metrics for evaluation? (\S \ref{sec:component}); 
(RQ3) How does the performance vary when a system is evaluated using tasks of different complexities, e.g., from single-domain to multi-domain tasks? (\S \ref{sec:domain}); 
(RQ4) Does simulated evaluation correlate well with human evaluation? (\S \ref{sec:human}). 

Our results show that 
(1) pipeline systems trained using fine-grained supervision signals at different component levels often achieve better overall performance than the joint models and end-to-end systems, 
(2) the results of component-wise, single-turn evaluation are not always consistent with that of system-wise, multi-turn evaluation, 
(3) as expected, the performance of dialog systems of all three types drops significantly with the increase of task complexity, and 
(4) despite the discrepancy between simulators and human users, simulated evaluation correlates moderately with human evaluation, indicating that simulated evaluation is still a valid alternative to the costly human evaluation, especially in the early stage of development.





\section{Experimental Setting}

\subsection{Data}
In order to conduct a system-wise evaluation and an in-depth empirical analysis of various dialog systems, we adopt the MultiWOZ \cite{budzianowski2018multiwoz} corpus in this paper. It is a multi-domain, multi-intent task-oriented dialog corpus that contains 3,406 single-domain dialogs and 7,032 multi-domain dialogs, with 13.18 tokens per turn and 13.68 turns per dialog on average. The dialog states and system dialog acts are fully annotated. The corpus also provides the domain ontology that defines all the entities and attributes in the external databases. We also use the augmented annotation of user dialog acts from \cite{lee2019convlab}.

\begin{figure}[!tb]
    \centering
    \includegraphics[width=\linewidth]{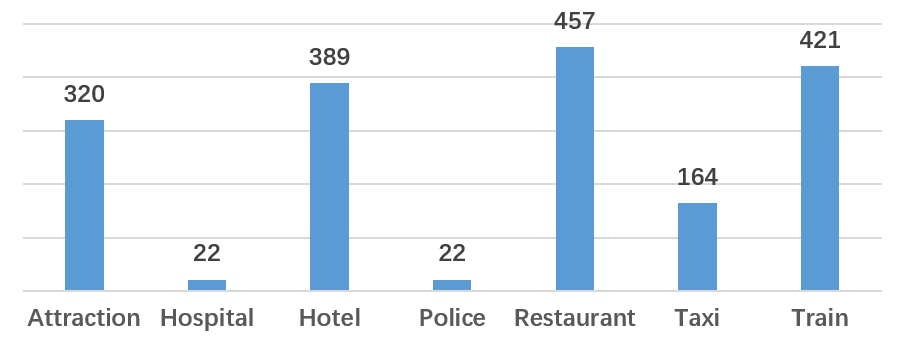}
    \caption{Domain distribution of the user goals used in the experiments. A goal with multiple domains is counted repeatedly for each domain.}
    \label{fig:goal}
\end{figure}

\subsection{User Goal}
During evaluation, a dialog system interacts with a simulated or 
human user to accomplish a task according to a pre-defined user goal. A user goal is the description of the state that a user wants to reach in a conversation, containing \textit{indicated constraints} (e.g., a restaurant serving Japanese food in the center of the city) and \textit{requested information} (e.g., the address, phone number of a restaurant). 

A user goal is initialized to launch the dialog session during evaluation. To ensure a fair comparison, we apply a fixed set of 1,000 user goals for both simulated and human evaluation. In the goal sampling process, we first obtain the frequency of each slot in the dataset and then sample a user goal from the slot distribution. We also apply additional rules to remove inappropriate combinations, e.g., a user cannot inform and inquire about the arrival time of a train in the same session. In the case where no matching database entry exists based on the sampled goal, we resample a new user goal until there is an entity in the database that satisfies the new constraints. In evaluation, the user first communicates with the system based on the initial constraints, and then can change the constraints if the system informs the user that the requested entity is not available. 
The detailed distribution of these goals is shown in Fig. \ref{fig:goal}. Among the 1,000 user goals, the numbers of goals involving 1/2/3 domains are 328/549/123, respectively.

\subsection{Platform and Simulator}
We use the open-source end-to-end dialog system platform, ConvLab \cite{lee2019convlab}, as our experimental platform. ConvLab enables researchers to develop a dialog system using preferred architectures and supports system-wise simulated evaluation. It also provides an integration of crowd-sourcing platforms such as Amazon Mechanical Turk for human evaluation.

\begin{figure}[!tb]
    \centering
    \includegraphics[width=0.85\linewidth]{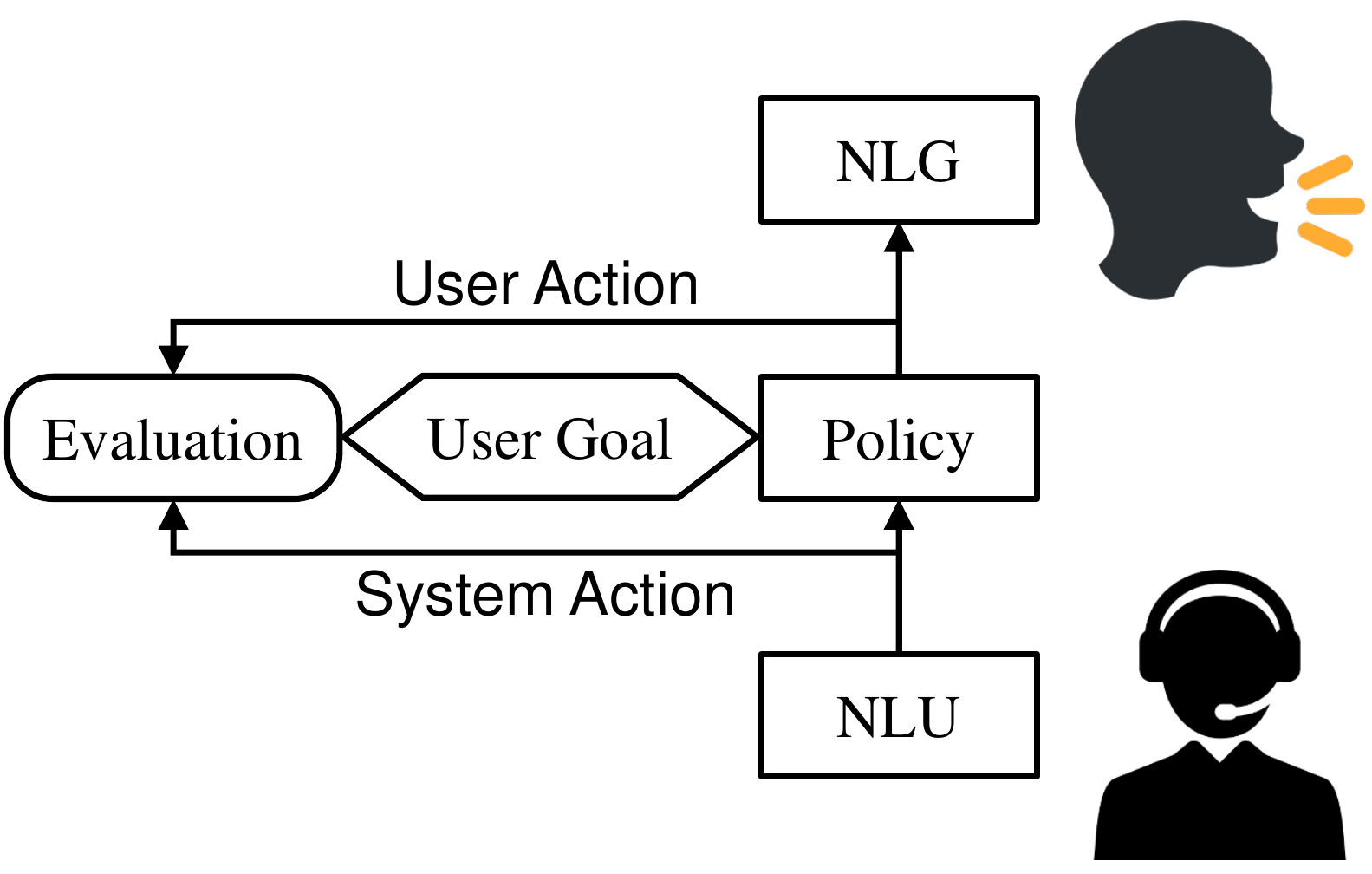}
    \caption{The framework of a user simulator and the mechanism for simulated evaluation.}
    \label{fig:user}
\end{figure}

To automatically evaluate a multi-turn dialog system, Convlab implements an 
agenda-based user simulator \cite{schatzmann2007agenda}. 
Given a user goal, the simulator's policy uses a stack-like structure with complex hand-crafted heuristics to inform its goal and mimics complex user behaviors during a conversation. Since the system interacts with the simulator in natural language, the user simulator directly takes system utterances as input and outputs a user response. The overall architecture of user simulator is presented in Fig. \ref{fig:user}. It consists of three modules: NLU, policy, and NLG. 
We use the default configuration of the simulator in Convlab: 
a RNN-based model MILU  
(\textbf{M}ulti-\textbf{I}ntent \textbf{L}anguage \textbf{U}nderstanding, extended \cite{hakkani2016multi}) for NLU,
a hand-crafted policy, 
and a retrieval model for NLG.

\subsection{Evaluation Metrics}
We use the number of \textit{dialog turns}, averaging over all dialog sessions, to measure the efficiency of accomplishing a task. 
A user utterance and a subsequent system utterance are regarded as one dialog turn. The system should help each user accomplish his/her goal within 20 turns, otherwise the dialog is regarded as failure.
We utilize two other metrics: \textit{inform F1} and \textit{match rate} to estimate the task success. Both metrics are calculated based on the \textit{dialog act} \cite{stolcke2000dialogue}, an abstract representation that extracts the semantic information of an utterance. The dialog act from the input and output of the user simulator's policy will be used to calculate two scores, as shown in Fig. \ref{fig:user}. 
Inform F1 evaluates whether all the information \textit{requests} are fulfilled, and match rate assesses whether the offered entity meets all the \textit{constraints} specified in a user goal. The dialog is marked as successful if and only if both inform recall and match rate are 1. 

\subsection{System Configurations}
To investigate how much system-wise and component-wise evaluations differ, we compare a set of dialog systems that are assembled using different state-of-the-art modules and settings in our experiments. 
The full list of these systems are shown in Table \ref{tab:e2e_auto}, which includes 4 pipeline systems (\textit{SYSTEM-1$\sim$4}), 10 joint-model systems (\textit{SYSTEM-5$\sim$13}) and 2 end-to-end systems (\textit{SYSTEM-15$\sim$16}). Note that some systems (e.g. SYSTEM-4, SYSTEM-10) generate \textit{delexicalized} responses where the slot values are replaced with their slot names. We convert these responses to natural language by filling the slot values based on dialog acts and/or database query results.

In what follows, we briefly introduce these modules and the corresponding models\footnote{All state-of-the-art models mentioned in this paper are based on the open-source code that is available and executable as of February 29, 2020.} used in our experiments. 
The component-wise evaluation results of these modules are shown in Table \ref{tab:component}. For published works, we train all the models using the open-source code with the training, validation and test split offered in MultiWOZ, and replicate the performance reported in the original papers or on the leaderboard.

\paragraph{NLU}
A \textit{natural language understanding} module identifies user intents and extracts associated information from users' raw utterances. 
We consider two approaches that can handle multi-intents as reference: 
a RNN-based model MILU 
which extends \cite{hakkani2016multi} and is fine-tuned on multiple domains, intents and slots; 
and a fine-tuned BERT model \cite{devlin2019bert}. 
Following the joint tagging scheme \cite{zheng2017joint}, the labels of intent detection and slot filling are annotated for domain classification during training. 
Both models use dialog history up to the last dialog turn as context. 
Note that there can be multiple intents or slots in one sentence, we calculate two F1 scores for intents and slots, respectively.

\paragraph{DST}
A \textit{dialog state tracker} encodes the extracted information as a compact set of dialog state that contains a set of informable slots and their corresponding values (user constraints), and a set of requested slots\footnote{Dialog state can include everything a system must know in order to make a decision about what to do next, e.g., DSTC2 corpus \cite{henderson2014second} contains \textit{search method} representing user intents in the dialog state, but only aforementioned items are taken into account as our experiments are conducted on MultiWOZ in this paper.}.
We have implemented a rule-based DST to update the slot values in the dialog state based on the output of NLU. We then compare four word-level DST: a multi-domain classifier MDBT \cite{ramadan2018large} 
which enumerates all possible candidate slots and values, SUMBT \cite{lee2019sumbt} 
that uses a BERT encoder and a slot-utterance matching architecture for classification, TRADE \cite{wu2019transferable} 
that shares knowledge among domains to directly generate slot values, and COMER \cite{ren2019scalable} 
which applies a hierarchical encoder-decoder model for state generation. 
We use two metrics for evaluation. The joint goal accuracy compares the predicted dialog states to the ground truth at each dialog turn, and the output is considered correct if and only if all the predicted values exactly match the ground truth. The slot accuracy individually compares each (domain, slot, value) triplet to its ground truth label. 

\paragraph{Policy}
A \textit{dialog policy} relies on the dialog state provided by DST to select a system action.
We compare two dialog policies: a hand-crafted policy, and a reinforcement learning policy GDPL \cite{takanobu2019guided} that jointly learns a reward function. 
We also include in our comparison three joint models, known as word-level policies, which combine the policy and the NLG module to produce natural language responses from dialog states. 
They are MDRG \cite{wen2017network} 
where an attention mechanism is conditioned on the dialog states, 
HDSA \cite{chen2019semantically} 
that decodes response from predicted hierarchical dialog acts, 
and LaRL \cite{zhao2019rethinking} 
which uses a latent action framework. 
We use BLEU score \cite{papineni2002bleu}, inform rate and task success rate as metrics for evaluation. 
Note that the inform rate and task success for evaluating policies are computed at the turn level, while the ones used in system-wise evaluation are computed at the dialog level. 

\paragraph{NLG}
A \textit{natural language generation} module generates a natural language response from a dialog act representation.
We experiment with two models: a retrieval-based model that samples a sentence randomly from the corpus using dialog acts, and a generation-based model SCLSTM \cite{wen2015semantically} 
which appends a sentence planning cell in RNN.
To evaluate the performance of NLG, we adopt BLEU score to evaluate the quality of the generated text, and slot error rate (SER) to measure whether the generated response contains missing or redundant slot values.

\paragraph{E2E}
An \textit{end-to-end} model takes user utterances as input and directly output system responses in natural language. 
We experiment with two models: TSCP \cite{lei2018sequicity} 
that uses belief spans to represent dialog states, and DAMD \cite{zhang2020task} 
that further uses action spans to represent dialog acts as additional information. For single-turn evaluation, BLEU, inform rate and success rate are provided.

\begin{table*}[!tb]
    \centering
    \begin{tabular}{c|cccc|cccccc}
    \hline
        \multirow{2}{*}{ID} & \multicolumn{4}{c|}{Configuration} & \multirow{2}{*}{Turn} & \multicolumn{3}{c}{Inform} & \multirow{2}{*}{Match} & \multirow{2}{*}{Succ.} \\
    \cline{2-5} \cline{7-9}
        & NLU & DST & Policy & NLG &  & Prec. & Rec. & F1 \\
    \hline
        1 & BERT & rule & rule & retrieval & 6.79 & 0.79 & 0.91 & 0.83 & 90.54 & 80.9 \\ 
        2 & MILU & rule & rule & retrieval & 7.24 & 0.76 & 0.88 & 0.80 & 87.93 & 77.6 \\ 
        3 & BERT & rule & GDPL & retrieval & 10.86 & 0.72 & 0.69 & 0.69 & 68.34 & 54.1 \\
        4 & BERT & rule & rule & SCLSTM & 13.38 & 0.64 & 0.58 & 0.58 & 51.41 & 43.0 \\ 
        5 & \multicolumn{2}{c}{MDBT}  & rule & retrieval & 16.55 & 0.47 & 0.35 & 0.37 & 39.76 & 18.8 \\ 
        6 & \multicolumn{2}{c}{SUMBT}  & rule & retrieval & 13.71 & 0.51 & 0.44 & 0.44 & 46.44 & 27.8 \\ 
        7 & \multicolumn{2}{c}{TRADE}  & rule & retrieval & 9.56 & 0.39 & 0.41 & 0.37 & 38.37 & 22.4 \\ 
        8 & \multicolumn{2}{c}{COMER}  & rule & retrieval & 16.79 & 0.30 & 0.28 & 0.28 & 29.06 & 17.3 \\ 
        9 & BERT & rule & \multicolumn{2}{c|}{MDRG}  & 17.90 & 0.35 & 0.34 & 0.32 & 29.07 & 19.2 \\ 
        10 & BERT & rule & \multicolumn{2}{c|}{HDSA}  & 15.91 & 0.47 & 0.62 & 0.50 & 39.21 & 34.3 \\ 
        11 & BERT & rule & \multicolumn{2}{c|}{LaRL}  & 13.08 & 0.40 & 0.68 & 0.48 & 68.95 & 47.7 \\ 
        12 & \multicolumn{2}{c}{SUMBT} & \multicolumn{2}{c|}{HDSA} & 18.67 & 0.27 & 0.32 & 0.26 & 14.78 & 13.7 \\ 
        13 & \multicolumn{2}{c}{SUMBT} & \multicolumn{2}{c|}{LaRL} & 13.92 & 0.36 & 0.64 & 0.44 & 57.63 & 40.4 \\ 
        14 & \multicolumn{2}{c}{TRADE} & \multicolumn{2}{c|}{LaRL} & 14.44 & 0.35 & 0.57 & 0.40 & 36.07 & 30.8 \\ 
        15 & \multicolumn{4}{c|}{TSCP}  & 18.20 & 0.37 & 0.32 & 0.31 & 13.68 & 11.8 \\ 
        16 & \multicolumn{4}{c|}{DAMD} & 11.27 & 0.64 & 0.69 & 0.64 & 59.67 & 48.5 \\
    \hline
    \end{tabular}
    \caption{System-wise simulated evaluation with different configurations and models. We use \textit{SYSTEM-$<$ID$>$} to represent the configuration's abbreviation throughout the paper.}
    \label{tab:e2e_auto}
\end{table*}

\section{Empirical Analysis}

\subsection{Performance under Different Settings (RQ1)}\label{sec:config}
We compare the performance of three types of systems, pipeline, joint-model and end-to-end.
Results in Table \ref{tab:e2e_auto} show that pipeline systems often achieve better overall performance than the joint models and end-to-end systems because using fine-grained labels at the component level can help pipeline systems improve the task success rate.

\begin{figure}[!tb]
    \centering
    \includegraphics[width=0.8\linewidth]{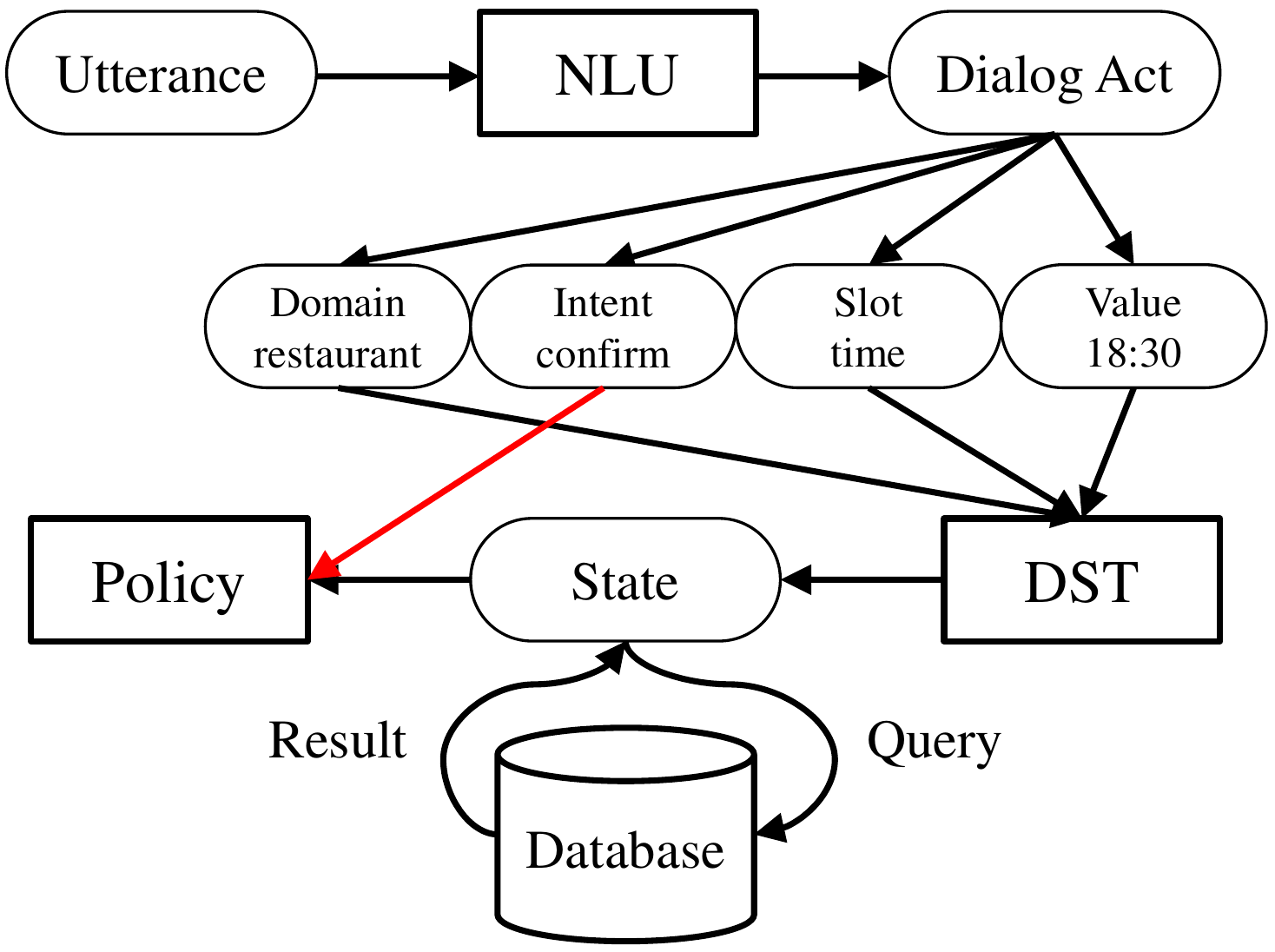}
    \caption{Illustration of NLU and DST in the dialog system. The intent information (red arrow) is missing in the dialog state on MultiWOZ if the system merges a word-level DST with a dialog policy.}
    \label{fig:NLU}
\end{figure}

\begin{figure}[!tb]
    \centering
    \includegraphics[width=\linewidth]{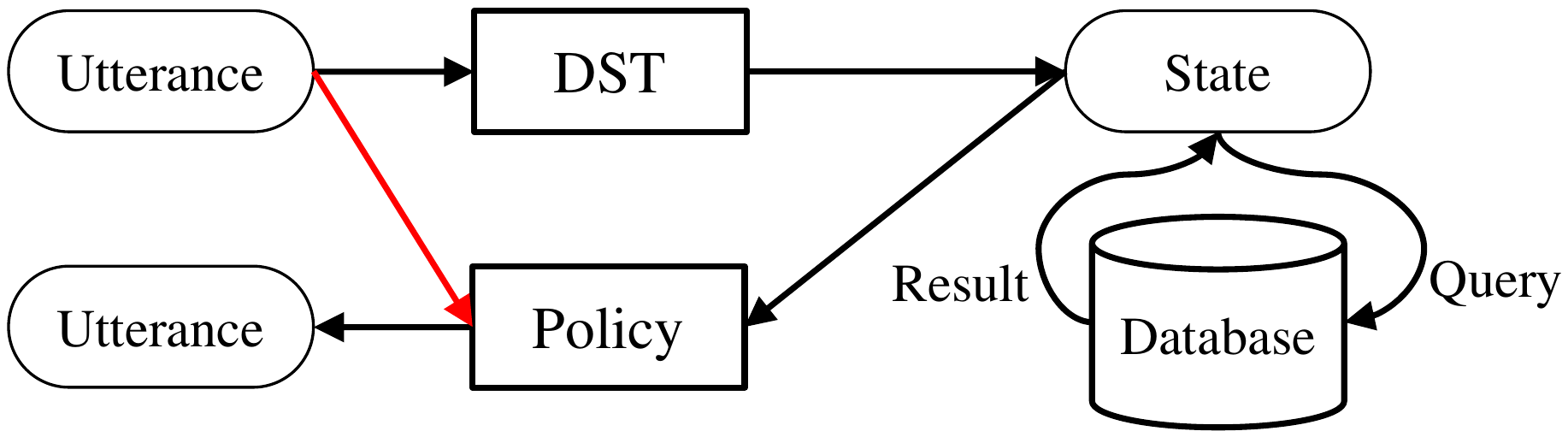}
    \caption{The common architecture of a system using word-level or end-to-end models. User utterances are encoded again (red arrow) for response generation.}
    \label{fig:unified}
\end{figure}

\paragraph{NLU with DST or joint DST}
It is essential to predict dialog states to determine what a user has expressed and wants to inquire. The dialog state is used to query the database, predict the system dialog act, and generate a dialog response. Although many studies have focused on the word-level DST that directly predicts the state using the user query, we also investigate the cascaded configuration where an NLU model is followed by a rule-based DST. As shown in Table \ref{tab:e2e_auto}, the success rate has a sharp decline when using word-level DST, compared to using an NLU model followed by a rule-based DST (17.3\%$\sim$27.8\% in \textit{SYSTEM-(5$\sim$8)} vs. 80.9\% in \textit{SYSTEM-1}). 
The main reason is that the dialog act predicted by NLU contains both slot-value pairs and \textbf{user intents}, whereas the dialog state predicted by the word-level DST only records the user constraints in the current turn, causing information loss for action selection (via dialog policy) as shown in Fig. \ref{fig:NLU}. For example, a user may want to confirm the booking time of the restaurant, but such an intent cannot be represented in the slot values. 
However, we can observe that word-level DST achieves better overall performance by combining with word-level policy, e.g., 40.4\% success rate in \textit{SYSTEM-13} vs. 27.8\% in \textit{SYSTEM-6}. This is because word-level policy implicitly detects user intents by encoding the \textbf{user utterance} as additional input, as presented in Fig. \ref{fig:unified}. Neverthsless, all those joint approaches still under-perform traditional pipeline systems.

\paragraph{NLG from dialog act or state}
We compare two strategies for generating responses. One is based on an ordinary NLG module that generates a response according to dialog act predicted by dialog policy. The other uses the word-level policy to directly generates a natural language response based on dialog state and user query. 
As we can see in Table \ref{tab:e2e_auto} that the performance drops substantially when we replace the retrieval NLG module with a joint model such as \textit{MDRG} or \textit{HDSA}. This indicates that the dialog act has encoded sufficient semantic information so that a simple retrieval NLG module can give high-quality replies. 
However, the fact, that \textit{SYSTEM-11} which uses word-level policy \textit{LaRL} even outperforms \textit{SYSTEM-4} which uses the NLG model \textit{SCLSTM} in task success (47.7\% vs. 43.0\%), indicates that response generation can be improved by jointly training policy and NLG modules. 

\paragraph{Database query}
As part of dialog management, it is crucial to identify the correct entity that satisfies the user goal. 
MultiWOZ contains a large number of entities across multiple domains, making it impossible to explicitly learn the representations of all the entities in the database as previous work did \cite{dhingra2017towards,madotto2018mem2seq}. This requires the designed system to deal with a large-scale external database, which is closer to reality. It can be seen in Table \ref{tab:e2e_auto} that most joint models have a lower match rate than the pipeline systems. 
In particular, \textit{SYSTEM-15} rarely selects an appropriate entity during the dialog (13.68\% match rate) since the proposed \textit{belief spans} only copy the values from utterances without knowing which domain or slot type the values belong to. Due to the poor performance in dialog state prediction, it cannot consider the external database selectively, thereby failing to satisfy the user's constraints. In comparison, \textit{SYSTEM-16} has achieved the highest success rate (48.5\%) and the second-highest match rate (59.67\%) among all the systems using joint models (\textit{SYSTEM-5$\sim$14}). This is because \textit{DAMD} utilizes \textit{action spans} to predict both user and system dialog acts in addition to belief spans, which behaves like a pipeline system. This indicates that an explicit dialog act supervision can improve dialog state tracking.



\begin{table}[!tb]
    \centering
    \begin{subtable}{\linewidth}
        \centering
        \begin{tabular}{c|cccc}
        \hline
            Model & Slot & Intent & Overall\\
        \hline
            MILU & 81.90 & 85.82 & 83.27 \\ 
            BERT & 84.25 & 89.84 & 86.21 \\ 
        \hline
        \end{tabular}
        \caption{NLU}
        \label{tab:nlu}
    \end{subtable}
    \begin{subtable}{\linewidth}
        \centering
        \begin{tabular}{c|cc}
        \hline
            Model & Slot Acc. & Joint Acc.\\
        \hline
            MDBT$^\dag$ & 89.53 & 15.57 \\
            SUMBT$^\dag$ & 96.44 & 46.65 \\
            TRADE$^\dag$ & 96.92 & 48.62 \\
            COMER & 95.52 & 48.79 \\
        \hline
        \end{tabular}
        \caption{Word-level DST}
        \label{tab:dst}
    \end{subtable}
    \begin{subtable}{\linewidth}
        \centering
        \begin{tabular}{c|ccc}
        \hline
            Model & BLEU & Inform & Succ.\\
        \hline
            MDRG$^\dag$ & 18.8 & 71.3 & 61.0 \\
            HDSA$^\dag$ & 23.6 & 82.9 & 68.9 \\
            LaRL$^\dag$ & 12.8 & 82.8 & 79.2 \\
        \hline
        \end{tabular}
        \caption{Word-level Policy}
        \label{tab:policy}
    \end{subtable}
    \begin{subtable}{\linewidth}
        \centering
        \begin{tabular}{c|cc}
        \hline
            Model & BLEU & SER\\
        \hline
            Retrieval & 33.1 & -- \\
            SCLSTM & 51.6 & 3.10 \\
        \hline
        \end{tabular}
        \caption{NLG}
        \label{tab:nlg}
    \end{subtable}
    \begin{subtable}{\linewidth}
        \centering
        \begin{tabular}{c|ccc}
        \hline
            Model & BLEU & Inform & Succ.\\
        \hline
            TSCP & 15.5 & 66.4 & 45.3 \\
            DAMD & 16.6 & 76.3 & 60.4 \\
        \hline
        \end{tabular}
        \caption{E2E}
        \label{tab:e2e}
    \end{subtable}
    \caption{Component-wise performance of each module. $\dag$: results from the MultiWOZ leaderboard.}
    \label{tab:component}
\end{table}

\begin{table*}[!tb]
    \centering
    \small
    \begin{tabular}{c|cccc|cccc|ccc}
        \hline
        \multirow{2}{*}{ID} & \multicolumn{4}{c|}{Restaurant} & \multicolumn{4}{c|}{Train} & \multicolumn{3}{c}{Attraction}\\
        \cline{2-12}
         & Turn & Info. & Match & Succ. & Turn & Info. & Match & Succ. & Turn & Info. & Succ. \\
        \hline
        1 & 2.82 & 0.94 & 96.9 & 98 & 3.06 & 1.0 & 100 & 100 & 3.12 & 0.69 & 63 \\
        2 & 2.84 & 0.92 & 100 & 98 & 2.99 & 1.0 & 94.2 & 97 & 3.70 & 0.73 & 65 \\
        3 & 8.68 & 0.70 & 69.4 & 70 & 6.07 & 0.80 & 67.3 & 75 & 5.61 & 0.67 & 62 \\
        4 & 6.00 & 0.77 & 68.8 & 78 & 11.53 & 0.71 & 67.3 & 55 & 12.57 & 0.57 & 46 \\
        6 & 9.41 & 0.64 & 72.7 & 60 & 5.13 & 0.97 & 90.4 & 93 & 14.79 & 0.23 & 9 \\ 
        11 & 9.91 & 0.39 & 66.7 & 61 & 4.02 & 0.86 & 88.5 & 97 & 4.73 & 0.68 & 80 \\
        13 & 8.35 & 0.40 & 65.6 & 60 & 4.19 & 0.85 & 94.2 & 96 & 6.06 & 0.60 & 73 \\
        15 & 14.72 & 0.37 & 11.5 & 27 & 16.02 & 0.46 & 11.5 & 25 & 16.12 & 0.51 & 24 \\
        16 & 6.36 & 0.80 & 92.2 & 90 & 10.21 & 0.61 & 55.8 & 58 & 8.32 & 0.69 & 67 \\
        \hline
    \end{tabular}
    \caption{Performance with different single domain. Most systems achieve better performance in \textit{Restaurant} and \textit{Train} than \textit{Attraction}.}
    \label{tab:dom_category}
\end{table*}
\begin{table*}[!tb]
    \centering
    \small
    \begin{tabular}{c|cccc|cccc|cccc}
        \hline
        \multirow{2}{*}{ID} & \multicolumn{4}{c|}{Single} & \multicolumn{4}{c|}{Two} & \multicolumn{4}{c}{Three}\\
        \cline{2-13}
         & Turn & Info. & Match & Succ. & Turn & Info. & Match & Succ. & Turn & Info. & Match & Succ. \\
        \hline
        1 & 3.22 & 0.84 & 84.7 & 87 & 6.96 & 0.81 & 94.9 & 78 & 8.15 & 0.82 & 88.4 & 69 \\
        2 & 3.90 & 0.78 & 79.7 & 82 & 6.74 & 0.76 & 95.3 & 72 & 10.54 & 0.79 & 85.0 & 66 \\
        3 & 9.18 & 0.67 & 66.7 & 60 & 12.38 & 0.60 & 42.9 & 42 & 13.55 & 0.50 & 44.6 & 21 \\
        4 & 8.65 & 0.66 & 58.3 & 62 & 17.24 & 0.38 & 28.0 & 14 & 18.03 & 0.46 & 24.4 & 13 \\
        6 & 10.35 & 0.44 & 60.4 & 41 & 14.74 & 0.44 & 50.9 & 17 & 15.97 & 0.25 & 20.9 & 0 \\ 
        11 & 8.79 & 0.45 & 72.2 & 55 & 13.37 & 0.52 & 74.0 & 59 & 19.30 & 0.39 & 50.4 & 0 \\
        13 & 8.48 & 0.45 & 62.5 & 61 & 14.08 & 0.45 & 61.0 & 47 & 18.95 & 0.36 & 40.7 & 0 \\
        15 & 15.09 & 0.33 & 10.0 & 26 & 19.10 & 0.25 & 17.8 & 8 & 20.00 & 0.19 & 0.0 & 1 \\
        16 & 8.89 & 0.66 & 68.1 & 65 & 13.48 & 0.52 & 57.1 & 34 & 18.59 & 0.58 & 45.5 & 12 \\
        \hline
    \end{tabular} 
    \caption{Performance with different number of domains. All systems have performance drop as the number of domains increases.}
    \label{tab:dom_number}
\end{table*}

\subsection{Component-wise vs. System-wise Evaluation (RQ2)}\label{sec:component}
It is important to verify whether the component-wise evaluation is consistent with system-wise evaluation. By comparing the results in Table \ref{tab:e2e_auto} and Table \ref{tab:component}, we can observe that sometimes they are consistent (e.g., \textit{BERT} $>$ \textit{MILU} in Table \ref{tab:nlu}, and \textit{SYSTEM-1} $>$ \textit{SYSTEM-2}), but not always (e.g., \textit{TRADE} $>$ \textit{SUMBT} in Table \ref{tab:dst}, but \textit{SYSTEM-6} $>$ \textit{SYSTEM-7}).

In general, a better NLU model leads to a better multi-turn conversation, and \textit{SYSTEM-1} outperforms all other configurations in completing user goals. 
With respect to DST, though word-level DST models directly predict dialog states without explicitly detecting user intents, most of them perform poorly in terms of joint accuracy as shown in Table \ref{tab:dst}. This severely harms the overall performance because the downstream tasks strongly rely on the predicted dialog states. Interestingly, \textit{TRADE} has higher accuracy than \textit{SUMBT} on DST. But \textit{TRADE} performs worse than \textit{SUMBT} in system-wise evaluation (22.4\% in \textit{SYSTEM-7} vs. 27.8\% in \textit{SYSTEM-6}). The observation is similar to \textit{COMER} vs. \textit{TRADE}. 
This indicates that the results of component-wise evaluation in DST are not consistent with those of system-wise evaluation, which may be attributed to the noisy dialog state annotations \cite{eric2019multiwoz}. 

As for word-level policy, \textit{HDSA} that uses explicit dialog acts in supervision has higher BLEU than \textit{LaRL} that uses latent dialog acts, but \textit{LaRL} that is finetuned with reinforcement learning has much higher match rate than \textit{HDSA} in system-wise evaluation (68.95\% vs. 39.21\%). Although there is small difference between \textit{MDRG} and \textit{HDSA} in component-wise evaluation (61.0\% vs. 68.9\% in Table \ref{tab:policy}), the gap is increased (19.2\% in \textit{SYSTEM-9} vs. 34.3\% in \textit{SYSTEM-10}) in system-wise evaluation. In addition, even \textit{SCLSTM} achieves a higher BLEU score than the retrieval-based model (51.6\% vs. 33.1\% in Table \ref{tab:nlg}), it only obtains a lower success rate (43.0\% in \textit{SYSTEM-4} vs. 80.9\% in \textit{SYSTEM-1}) when assembled with other modules.
These results show again the discrepancy between component-wise and system-wise evaluation. 
The superiority of the systems using retrieval models may imply that lower SER in NLG is more critical than higher BLEU in goal-oriented dialog systems.

\paragraph{Error in multi-turn interactions}
Most existing work only evaluates the model with single-turn interactions. For instance, \textit{inform rate} and \textit{task success} at each dialog turn are computed given the current user utterance, dialog state and database query results for context-to-context generation \cite{wen2017network,budzianowski2019hello}. A strong assumption is that the model would be fed with the ground truth from the upstream modules or the last dialog turn. However, this assumption does not hold since a goal-oriented dialog consists of a sequence of associated inquiries and responses between the system and its user, and the system may produce erroneous output at any time. The errors may propagate to the downstream modules and affect the following turns. For instance, end-to-end models get worse success rate in multi-turn interactions than in single-turn evaluation in Table \ref{tab:e2e}. A sample dialog from \textit{SYSTEM-1} and \textit{SYSTEM-6} is provided in Table \ref{tab:sample_auto}. \textit{SYSTEM-6} does not extract the \textit{pricerange} slot (highlighted in red color) correctly. The incorrect dialog state further harms the performance of dialog policy, and the conversation gets stuck where the user (simulator) is always asking for the postcode, thereby failing to complete the task.

To summarize, the component-wise, single-turn evaluation results do not reflect the real performance of the system well, and it is essential to evaluate a dialog system in an end-to-end, interactive setting. 


\subsection{Performance of Task with Different Complexities (RQ3)}\label{sec:domain}
With the increasing demands to address various situations in multi-domain dialog, we choose 9 representative systems across different configurations and approaches 
to further investigate how their performance varies with the complexities of the tasks. 100 user goals are randomly sampled under each domain setting. 
Results in Table \ref{tab:dom_category} and \ref{tab:dom_number} show that the overall performance of all systems varies with different task domains and drops significantly with the increase of task complexity, while pipeline systems are relatively robust to task complexity. 

\paragraph{Performance with different single domains}
Table \ref{tab:dom_category} shows the performance with respect to different single domains.
\textit{Restaurant} is a common domain where users inquiry some information about a restaurant and make reservations. \textit{Train} has more entities and its domain constraints can be more complex, e.g., the preferred train should \textit{arrive before} 5 p.m. \textit{Attraction} is an easier one where users do not make reservations. There are 7/6/3 informable slots that need to be tracked in \textit{Restaurant}/\textit{Train}/\textit{Attraction} 
respectively. Surprisingly, most systems perform better in \textit{Restaurant} or \textit{Train} than \textit{Attraction}. 
This may result from the noise database in \textit{Attraction} where \textit{pricerange} information is missing sometimes, and from the uneven data distribution where \textit{Restaurant} and \textit{Train} appear more frequently in the training set. 
In general, pipeline systems perform more stably across multiple domains 
than joint models and end-to-end systems.

\paragraph{Performance with different number of domains}
Table \ref{tab:dom_number} demonstrates how the performance varies with the number of domains in a task. We can observe that most systems fall short to deal with multi-domain tasks. Though some systems such as \textit{SYSTEM-13} and \textit{SYSTEM-16} can achieve a relatively high inform F1 or match rate for a single domain, the overall success rate drops substantially on two-domain tasks, and most systems fail to complete three-domain tasks. The number of dialog turns also increases remarkably when the number of domains increases. Among all these configurations, only the pipeline systems \textit{SYSTEM-2} and \textit{SYSTEM-1} can keep a high success rate when there are three domains in a task. These results show that current dialog systems are still insufficient to deal with complex tasks, and that pipeline systems outperform 
joint models and end-to-end systems.

\subsection{Simulated vs. Human Evaluation (RQ4)}\label{sec:human}
Since the ultimate goal of a task-oriented dialog system is to help users accomplish real-world tasks, it is essential to justify the correlation between simulated and human evaluation. 
For human evaluation, 100 Amazon Mechanical Turk workers are hired to interact with each system and then give their judgement on task success. The ability of Language Understanding (LU) and Response Appropriateness (RA) of the systems are assessed at the same time, and each worker gives a score on these two metrics with a five-point scale. We compare 5 systems that achieve the best performance in the simulated evaluation under different settings. 

\begin{table}[!tb]
    \centering
    \begin{tabular}{c|cccc|c}
    \hline
        ID & Turn & LU & RA & Succ. & Corr. \\
    \hline
        1 & 18.58 & 3.62 & 3.69 & 62 & 0.57\\ 
        6 & 20.63 & 2.85 & 2.91 & 27 & 0.72\\ 
        11 & 19.98 & 2.36 & 2.41 & 23 & 0.53\\ 
        13 & 19.26 & 2.17 & 2.49 & 14 & 0.46\\ 
        16 & 16.33 & 2.61 & 2.65 & 23 & 0.55\\
    \hline
    \end{tabular}
    \caption{System-wise evaluation with human users. Correlation coefficient between simulated and human evaluation is presented in the last column.} 
    \label{tab:e2e_human}
\end{table}

Table \ref{tab:e2e_human} shows the human evaluation results of 5 dialog systems. Comparing with the simulated evaluation in Table \ref{tab:e2e_auto}, we can see that Pearson's correlation coefficient lies around 0.5 to 0.6 for most systems, indicating that simulated evaluation correlates moderately well with human evaluation. Similar to simulated evaluation, the pipeline system \textit{SYSTEM-1} obtains the highest task success rate in human evaluation. A sample human-machine dialog from \textit{SYSTEM-1} and \textit{SYSTEM-6} is provided in Table \ref{tab:sample_human}. The result is similar to the simulated session in Table \ref{tab:sample_auto} but \textit{SYSTEM-6} fails to respond with the \textit{phone} number in Table \ref{tab:sample_human} instead (highlighted in red color). All these imply the reliability of the simulated evaluation in goal-oriented dialog systems, showing that simulated evaluation can be a valid alternative to the costly human evaluation for system developers.

However, compared to simulated evaluation, we can observe that humans converse more naturally than the simulator, e.g., the user confirms with \textit{SYSTEM-1} whether it has booked 7 seats in Table \ref{tab:sample_human}, and most systems have worse performance in human evaluation. This indicates that there is still a gap between simulated and human evaluation. This is due to the discrepancy between the corpus and human conversations. The dataset only contains limited human dialog data, on which the user simulator is built. Both the system and the simulator are hence limited by the training corpus. As a result, the task success rate of most systems decreases significantly in human evaluation, e.g., from 40.4\% to 14\% in \textit{SYSTEM-13}. This indicates that existing dialog systems are vulnerable to the variation of human language (e.g., the sentence highlighted in brown in Table \ref{tab:sample_human}), which demonstrates a lack of robustness in dealing with real human conversations.

\section{Related Work} 

Developers have been facing many problems when evaluating a goal-oriented dialog system. A range of well-defined automatic metrics have been designed for different components in the system, e.g., joint goal accuracy in DST and task success rate in policy optimization introduced in Table \ref{tab:dst} and \ref{tab:policy}. A broadly accepted evaluation scheme for the goal-oriented dialog was first proposed by PARADISE \cite{walker1997paradise}. It estimates the user satisfaction by measuring two types of aspects, namely \textit{dialog cost} and \textit{task success}. \citesub{paek2001empirical} suggests that a useful dialog metric should provide an estimate of how well the goal is met and allow for a comparative judgement of different systems. Though a model can be optimized against these metrics via supervised learning, each component is trained or evaluated separately, thus difficult to reflect real user satisfaction. 

As human evaluation by asking crowd-sourcing workers to interact with a dialog system is much expensive \cite{ultes2013quality,su2016line} and prone to be affected by subjective factors \cite{higashinaka2010issues,schmitt2015interaction}, researchers have tried to realize automatic evaluation of dialog systems. Simulated evaluation \cite{araki1996automatic,eckert1997user} is widely used in recent works \cite{williams2017hybrid,peng2017composite,takanobu2019guided,takanobu2020multi} and platforms \cite{ultes2017pydial,lee2019convlab,papangelis2020plato,zhu2020convlab2}, where the system interacts with a user simulator which mimics human behaviors. Such evaluation can be conducted at the dialog act or natural language level. The advantages of using simulated evaluation are that it can support multi-turn language interaction in a full end-to-end fashion and generate dialogs unseen in the original corpus. 

\section{Conclusion and Discussion}


In this paper, we have presented the system-wise evaluation result and empirical analysis to estimate the practicality of goal-oriented dialog systems with a number of configurations and approaches. Though our experiments are only conducted on MultiWOZ, we believe that such results can be generalized to all goal-oriented scenarios in dialog systems. We have the following observations:

1) We find that rule-based pipeline systems generally outperform state-of-the-art joint systems and end-to-end systems, in terms of both overall performance and robustness to task complexity.
The main reason is that fine-grained supervision on dialog acts would remarkably help the system plan and make decisions, because the system should predict the user intent and take proper actions during the conversation. This supports that good pragmatic parsing (e.g. dialog acts) is essential to build a dialog system.

2) Results show that component-wise, single-turn evaluation results are not always consistent with the overall performance of dialog systems. In order to accurately assess the effectiveness of each module, system-wise, multi-turn evaluation should be used from the practical perspective.
We advocate assembling the proposed model of a specific module into a complete system, and evaluating the system with simulated or human users via a standardized dialog platform, such as Rasa \cite{bocklisch2017rasa} or ConvLab.
Undoubtedly, this will realize a full assessment of the module's contribution to the overall performance, and facilitate fair comparison with other approaches.

3) Simulated evaluation can have a good assessment of goal-oriented dialog systems and show a moderate correlation with human evaluation, but it remarkably overestimates the system performance in human interactions. Thus, there is a need to devise better user simulators that resemble humans more closely. A simulator should be able to generate a natural and diverse response, and may change goals in complex dialog, etc. In addition, the simulator itself may make mistakes which derive the wrong estimation of the performance. However even with human evaluation a dialog system needs to deal with more complicated and uncertain situations. Therefore, it is vital to enhance the robustness of the dialog systems. Despite the discrepancy between simulators and human users, simulated evaluation is still a valid alternative to the costly human evaluation especially in the early stage of development.



\section*{Acknowledgements}
This work was jointly supported by the NSFC projects (Key project with No. 61936010 and regular project with No. 61876096), and the National Key R\&D Program of China (Grant No. 2018YFC0830200). We would like to thank anonymous reviewers for their valuable suggestions, and Sungjin Lee for helpful discussions.





\bibliography{acl2020}
\bibliographystyle{acl_natbib}

\appendix

\begin{table*}[!tb] 
\begin{tabular}{c|c|c|c|c}
\hline
\multicolumn{3}{c|}{\textbf{Hotel}} & \multicolumn{2}{c}{\textbf{Restaurant}} \\
\hline
Info & Reqt & Book & Info & Book \\
\hline
parking=yes & postcode & stay=2 & food=british & day=sunday \\
pricerange=moderate & address & day=sunday & pricerange=moderate & time=16:30 \\
& & & area=centre & people=8  \\
\hline
\end{tabular}
\begin{tabular}{l@{~}p{7cm}|l@{~}p{7cm}}
\hline
\multicolumn{2}{c|}{\textbf{SYSTEM-1}}  & \multicolumn{2}{c}{\textbf{SYSTEM-6}} \\
\hline
U:& I would prefer it in the centre area . Actually I need a moderately priced restaurant . Are there any fitting that description ? This restaurant should serve british food too . & U:& I would prefer it in the centre area . Actually I need a moderately priced restaurant . Are there any fitting that description ? This restaurant should serve british food too .  \\
D:& \textit{restaurant-food:british, restaurant-pricerange:moderate, restaurant-area:centre} &D:& \textit{restaurant-food:british, restaurant-pricerange:moderate, restaurant-area:centre} \\
S:& I have 4 options for you ! How about the copper kettle ? It is located in the centre . & S:& I have 4 options for you ! It is in the centre area . That is a british restaurant . the copper kettle is just your thing . It is in the centre area .  \\
U:& I need a table for 8 people . Can you try to book it at 16:30 . We will be there sunday night . & U:& It will be for 8 people . Can you try for 16:30 ? I am also looking for a restaurant to book for sunday.  \\
D:& \textit{restaurant-people:8, restaurant-day:sunday, restaurant-time:16:30} &D:& \textit{restaurant-people:8, restaurant-day:sunday, restaurant-time:16:30} \\
S:& Your reference number is 00000013 . & S:& All set . Your reference number is 00000013 . \\
U:& I would like to keep it in the moderate range , please . Oh , i need parking. & U:& Hello , I would like to find a hotel that includes free parking . I also need a moderate place to stay . \\
D:& \textcolor{blue}{\textit{hotel-parking:yes, hotel-pricerange:moderate}} &D:& \textcolor{red}{\textit{hotel-parking:yes}} \\
S:& \textcolor{blue}{I have 14 different options for you ! How about hamilton lodge ? Fits your request perfectly . It has free wifi .} & S:& \textcolor{red}{Reference number is : 00000013 .} \\
U:& That sounds fine , I just need the postcode though . &  U:& What is the postcode ? \\
D:& &D:& \\
S:& The postal code for that hotel is cb41da . &  S:& Thank you for using our services .  \\
U:& I need a room starting on sunday . There is going to be 8 of us. How about only 2 nights . & U:& That sounds fine , I just need the postcode though . \\
D:& \textit{hotel-day:sunday, hotel-people:8, hotel-stay:2} &D:& \\
S:& Here is the booking information : Booking was successful . Reference number is : 00000001. & S:& Glad to have been of help . Thank you for using the service . Enjoy the rest of your day ! \\
U:& Thank you very much , goodbye . & U:& I just need the postcode . \\
D:& &D:& \\
S:& You are very welcome. Goodbye. & S:& Ok , thank you . Have a good day . \\
\hline
\multicolumn{2}{c|}{\textbf{Success}} & \multicolumn{2}{c}{\multirow[t]{2}{*}[1.3ex]{\textbf{Failure:}} \begin{tabular}[c]{@{}p{6.5cm}@{}}System fails to track the value of price range in the dialog state and does not answer the postcode.\end{tabular}}
\\ \hline
\end{tabular}
\centering
\caption{A sample dialog session comparison between \textit{SYSTEM-1} and \textit{SYSTEM-6} in simulated evaluation. For simplicity, we only present the updated slots and values at each turn, instead of the entire dialog state (lines start with D).}
\label{tab:sample_auto}
\end{table*}

\begin{table*}[!tb]
\begin{tabular}{c|c|c|c|c}
\hline
\multicolumn{3}{c|}{\textbf{Train}} & \multicolumn{2}{c}{\textbf{Attraction}} \\
\hline
Info & Reqt & Book & Info & Reqt \\
\hline
destination=peterborough & duration & people=7 & area=centre & phone\\
day=saturday & trainID &  & & postcode \\
departure=cambridge & &  & & \\
leaveAt=13:00 & &  & &\\
\hline
\end{tabular}
\begin{tabular}{l@{~}p{7cm}|l@{~}p{7cm}}
\hline
\multicolumn{2}{c|}{\textbf{SYSTEM-1}}  & \multicolumn{2}{c}{\textbf{SYSTEM-6}} \\
\hline
U:& I am looking for a train that leaves at 13:00 & U:& Hi! I'm looking for a train departing from Cambridge after 13:00 \\
S:& What day would you like ? Where would you like your train to take you ? Where did you want to depart from ? & S:& What day did you have in mind ? Where are you heading to ? \\
U:& I would like to leave on Saturday and depart from Cambridge & U:& Saturday and heading to peterborough \\
S:& What is your destination ?  & S:& There is a train arriving at 13:00 would you like me to book tickets for that one ? I can book you on saturday. Would you like me to book a train to peterborough for you ? Would you like me to book a train from cambridge for you ? \\
U:& I need to go to Peterborough & U:& Yes, please book for 7 people \\
S:& We can book you for the train leaving at 13:00 . I can book you on Saturday. Would you like me to book a train to peterborough for you ? Would you like me to book a train from cambridge for you ? & S:& All set . Your reference number is 00000066 .  \\
U:& Yes. I need the train ID and booking for 7 people & U:& Thanks! I'm looking forward to seeing local tourist attractions \\
S:& Reference number is : 00000054 . &  S:& Thank you for using our services \\
U:& I want to see the tourist attractions so it should be near the Centre & U:& Do you know of any attractions in the centre?  \\
S:& There are 44 . Would you like me to recommend one for you ? Would you like vue cinema ? & S:& We have 44 options , can i recommend for you ? I would suggest sheep's green and lammas land park fen causeway . \\
U:& Yes. I also want the phone number and postcode for the cinema &  U:& \textcolor{brown}{Can I get the phone number for Sheep's Green?} \\
S:& \textcolor{blue}{The phone number is 08712240240 and postcode is cb11ps . They are located within the centre .} &  S:& \textcolor{red}{What time would you like to leave ?}  \\
U:& Do you have me booked for a party of 7 people? & U:& Around noon \\
S:& Here is the booking information : Booking was successful . Reference number is : 00000054. & S:& Have a good day . \\
U:& Are there 7 seats booked? & U:& Can you give me the phone number of the tourist attraction? \\
S:& Reference number is : 00000054 . & S:& \textcolor{red}{Thank you for contacting the help desk . Have a great day .} \\
U:& Thank You for your help and Goodbye! & U:& Can you give me a list of attractions?\\
S:& You are welcome . Have a good day ! & S:& Have a good day . \\
\hline
\multicolumn{2}{c|}{\textbf{Success}} & \multicolumn{2}{c}{\multirow[t]{2}{*}[1.3ex]{\textbf{Failure:}} \begin{tabular}[c]{@{}p{6.5cm}@{}}System does not answer the phone number of the requested tourist attraction.\end{tabular}} 
\\ \hline
\end{tabular}
\centering
\caption{A sample dialog session comparison between \textit{SYSTEM-1} and \textit{SYSTEM-6} in human evaluation.}
\label{tab:sample_human}
\end{table*}

\end{document}